\documentclass[letterpaper, 10 pt, conference]{ieeeconf} 
\usepackage{subcaption}
\usepackage{graphicx}
\usepackage{changepage}
\usepackage{booktabs}
\usepackage{balance}
\usepackage{color, colortbl}
\definecolor{Gray}{gray}{0.9}

\IEEEoverridecommandlockouts 

\usepackage{amsmath} 

\title{\LARGE \bf
MARLander: A Local Path Planning for Drone Swarms using Multiagent Deep Reinforcement Learning 
}

\author{Demetros Aschu, Robinroy Peter, Sausar Karaf, Aleksey Fedoseev, and Dzmitry Tsetserukou%
\thanks{The authors are with the Intelligent Space Robotics Laboratory, Skolkovo Institute of Science and Technology, Bolshoy Boulevard 30, bld. 1, 121205, Moscow, Russia
\tt \{Demetros.Aschu,Robinroy.Peter,Sausar Karaf,Aleksey.Fedoseev, D.Tsetserukou\}@skoltech.ru}} 

\begin{document}
\maketitle
\thispagestyle{empty}
\pagestyle{empty}


\begin{abstract}
Achieving safe and precise landings for a swarm of drones poses a significant challenge, primarily attributed to conventional control and planning methods. This paper presents the implementation of multi-agent deep reinforcement learning (MADRL) techniques for the precise landing of a drone swarm at relocated target locations. The system is trained in a realistic simulated environment with a maximum velocity of 3 m/s in training spaces of 4 x 4 x 4 m and deployed utilizing Crazyflie drones with a Vicon indoor localization system.

The experimental results revealed that the proposed approach achieved a landing accuracy of 2.26 cm on stationary and 3.93 cm on moving platforms surpassing a baseline method used with a Proportional–integral–derivative (PID) controller with an Artificial Potential Field (APF). This research highlights drone landing technologies that eliminate the need for analytical centralized systems, potentially offering scalability and revolutionizing applications in logistics, safety, and rescue missions.
\end{abstract}

\section{Introduction}
Swarm drones, characterized by their collaborative behavior, are driving research due to their disruptive potential across industries like agriculture, construction, entertainment, and logistics~\cite{a0, a1}. Challenges persist in achieving accurate landings on specified targets, highlighting the importance of safe descent to prevent damage and ensure mission success so addressing this challenge necessitates effective control and planning strategies.

A conventional approach of control and planning is susceptible to single points of failure, scalability issues, and high communication overhead due to routing decisions through a central hub, leading to delays and congestion \cite{a4}. Classical planners relying on full-state information face challenges with high-dimensional problems, demanding heavy computational resources \cite{a2}. Kinodynamic planners must accurately model drone dynamics near physical limits, impacting practical reliability \cite{a2, b15}. Conversely, conventional controllers like PID controllers are tuning-dependent and find it challenging to handle constraints. 
\begin{figure}[ht]
\centering
\includegraphics[width=0.48\textwidth ]{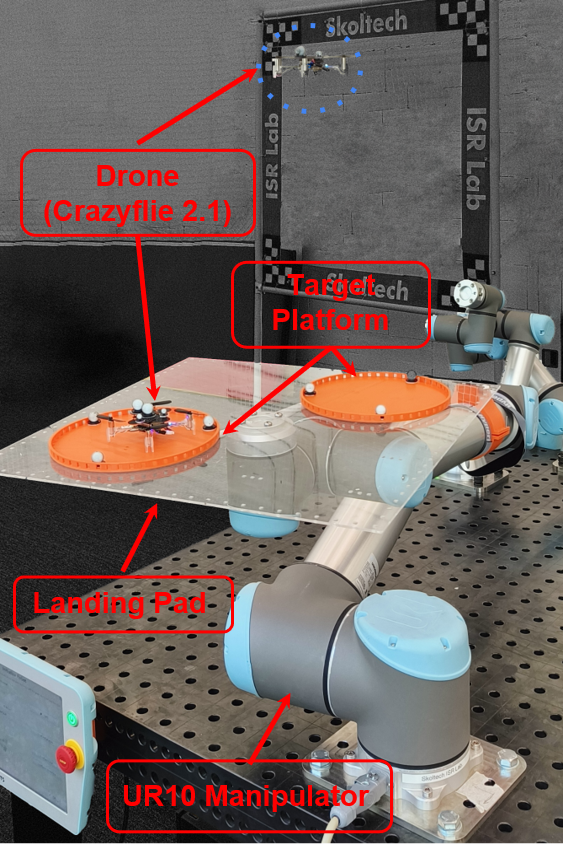}
\caption{ MARlander two drones landing on
the target platform placed on the robot manipulator}
 \label{fig:test}
\end{figure}
Moreover, model predictive controller (MPC) encounters issues, e.g., computational complexity and model dependency, which make them less robust in environments with complex dynamics and uncertainties \cite{a3, a23}.

In light of these challenges, deep reinforcement learning (DRL) has emerged as a critical tool in robotic applications, particularly for drones, enabling them to learn policies directly from raw data for informed decision-making based on their observation \cite{a5}. Extensive research highlights the effectiveness of DRL in single drone control, setting the stage for its expansion to multiple drones through a means of analytical centralization \cite{a3,b12}. However, the approach still inherits some challenges when working with multiple agents.

Significant advancements in multi-agent reinforcement learning (MARL) have driven the development of autonomous drone swarms capable of coordinated formations \cite{b13}, cooperative navigation~\cite{b14}, and path finding~\cite{a6, a7} that has been notably influential in many fields~\cite {a5}. However, implementing a multiagent reinforcement learning (MARL) approach, especially in maintaining individual agent behavior while fostering cooperative behavior, poses a significant challenge.

A centralized training approach, particularly which utilizing a single policy to govern interactions and decisions among multiple agents, has remained a key focus in current research~\cite{a8}, while another approach in recent research has pivoted towards decentralized multiagent systems where the individual agents can execute their control policy independently after training with supervised centralized system~\cite{b15, b16}.

This study introduces a multi-agent deep reinforcement learning (MADRL) approach for landing drone swarms in three-dimensional environments. The proposed solution utilizes a trained RL agent that receives local observations from each drone, enabling it to learn a control policy for the autonomous execution of landing trajectories. Preliminary results demonstrate the efficacy of the proposed approach in addressing the challenge of landing due to conventional approaches.

In our paper, we demonstrated the capability of our agent to land on the target platform with an acceptable accuracy of mean landing error of 2.26 cm, while also ensuring the avoidance of internal collisions during landing. This will be a base for the advancement in the landing of large fleets of drones in a decentralized manner, effectively overcoming the current challenges associated with it.

\section{Related Works}
Researchers have been addressing the challenge of developing an autonomous quadrotor capable of landing on statics platform~\cite{b0} and landing on dynamic platforms in turbulent wind conditions~\cite{b1} by developing a fully autonomous vision-based system. This system integrates localization, planning, and control components effectively. 

Moreover, multiple drone landing scenarios have been explored in truncation single agent drone landing showcasing advancements in control and planning methods such research work~\cite{b2} and~\cite{b3} show cooperative multiple drone landing and grasping task, Furthermore, the research~\cite{b4} shows Manipulator assisted landing where uses a tether and robot manipulation to enable multiple UAV landings without needing a large platform. The system incorporates MPC for stability and tracking, along with an adaptive estimator to handle motion disturbances.

The authors~\cite{b5} proposed a novel scientific method to achieve the dynamic landing of a heterogeneous swarm of drones on a moving platform. This method involves a leader drone equipped with a camera guiding follower drones through commands while ensuring collision avoidance via an APF, they also propose a system for landing on a moving platform using multiple agents under supervision~\cite{b16}.

A new approach to landing drones by incorporating neural networks into control and planning processes proposed by authors~\cite{b6} with a deep-learning-based robust nonlinear controller called the Neural-Lander, Furthermore, they demonstrated Neural-Swarm2 for controlling multi-rotor drone swarms~\cite{b7}.

However, due to a lack of robustness approach a single-agent autonomous landing has been enhanced using RL techniques the research work~\cite{b8} introduces an RL-based controller employing Least Square Policy Iteration (LSPI) to learn optimal control policies for generating landing trajectories.~\cite{b8} shows landing of a quadrotor on a platform, Additionally,~\cite{b7} suggested the autonomous landing of quadrotors on inclined surfaces, while~\cite{b8, b9} showcased Autonomous Landing on a platform with different vehicles and velocities. 

The proposed solution from~\cite{b12} introduced a method involving a single-agent RL-based control policy integrated with analytical planners such as APF to address collision avoidance within a drone swarm, with a trained RL controller directing drones toward a moving target for landing.

Despite improving upon the limitations of conventional controllers, this method encounters challenges when scaling to a large number of drones and managing complexity due to centralization.

In recent research, the MARL approach has been increasingly applied in drone swarm studies. For instance, authors~\cite{b15} developed a decentralized control policy allowing drones to map observations directly to motor thrusts, aiding in formation adjustments in 3D space. Additionally, in a study by~\cite{b13}, cooperative agents maintained a desired formation relative to two tracked agents while progressing towards a common objective. Furthermore, research by~\cite{b14} illustrated a MARL method for the collaborative navigation of Unmanned Aerial Vehicles (UAVs) using Centralized training Decentralized execution (CTDE).

\section{Methodology}
This section details the methodology used to develop and evaluate Lander, a multi-agent system enabling autonomous drones to land on target platforms. We employ the MADRL framework in a decentralized approach, which is described in detail below.
\subsection{System Overview}
\begin{figure}[h]
 \centering
 \includegraphics[width=0.5\textwidth]{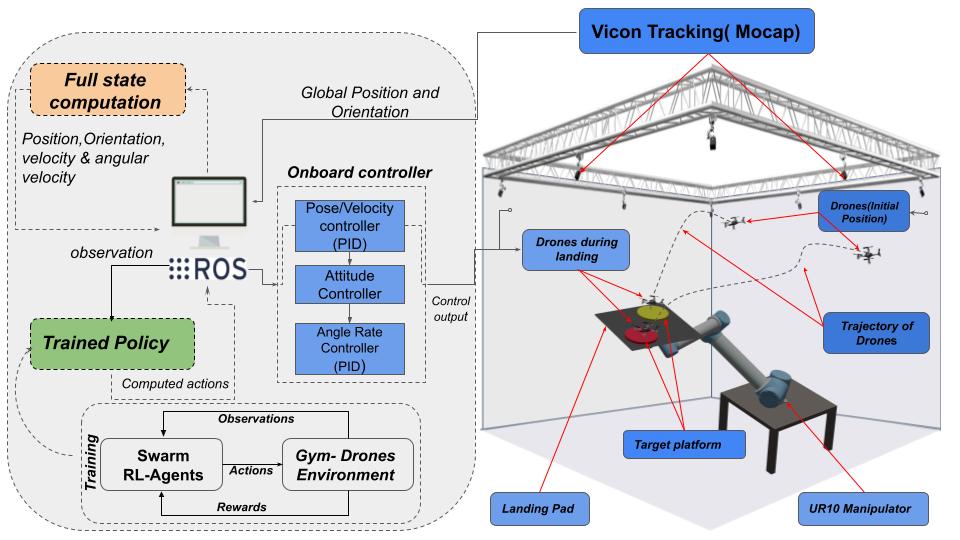}
 \caption{General system overview of MARLander.}
 \label{fig:system}
\end{figure}

The MARLander system depicted in Fig. \ref{fig:system} comprises a swarm of two Bitcraze Crazyflie drones, with a Universal Robot (UR10) affixed to the landing pad. Two separate target platforms are positioned on the landing pad, each spaced 0.5cm apart.

A VICON Vantage V5 motion capture system is utilized for micro drone localization, transmitting updated position and orientation data via the Robot Operating System (ROS) to the control station. The control station processes this information to estimate linear and angular velocity, forwarding it to a DRL model for policy generation. 

The policy generates control outputs at each time step, subsequently relayed to the Crazyflie onboard controller for motor actuation which generates trajectories for each drone, guiding them to land on their respective targets successfully.

\subsection{Problem Formulation and Preliminaries}
The landing problem is addressed using state-of-the-art DRL techniques. In a 3D environment with $N$ quadrotor drones, we utilize a Markov Decision Process (MDP) model denoted as: 
\begin{equation}
    MDP(O_t, A_t, R_t)
\end{equation}
where $O_t$ is the observation of the environment, consisting of $s_t$ which is the drone's state including kinematic information, and $g_t$ is the state of the target platform. $A_t$ is the actions, and $R_t$ is the reward signal guiding our objective.

The aim is to develop an optimal policy to map observations from each drone to control actions that drive it to land at the target position while preventing collisions among them. The optimal policy is formulated by maximizing the sum of discounted rewards.

\subsection{Environment Setup}

Our environment for training and assessing swarm landing in static and dynamic conditions is crafted to represent the MDP framework.

\subsubsection{Observation and Actions}
The observation vector of the $i^{th}$ drone in the swarm is denoted by:
\begin{equation} 
    {O_t}^i = [{p_t}^i, {q_t}^i, {v_t}^i, {\omega_t}^i]
\end{equation} 
where ${p_t}^i$ represents the drone's position relative to the goal point, ${q_t}^i$ denotes orientation, ${v_t}^i$ signifies a relative velocity and ${\omega_t}^i$ indicates angular velocity. Similarly, the action for the $i^{th}$ drone in the swarm is denoted by: 
\begin{equation}
    {A_t}^i = [{u_{x,t}}^i, {u_{y,t}}^i, {u_{z,t}}^i, ]
\end{equation}
where  $u_{,t}^i$ denotes the velocity of the $i^{th}$ drone within the swarm in each 3D axis. This velocity is subsequently utilized to compute the drone's next position as it progresses towards the target.

\subsubsection{Reward Function}
The reward signal denoted $ {R_t}^i$ is defined as follows:

\begin{equation} 
{R_t} =  \sum_{i=1}^{n}{{r_t}^{i}}+ r^{c} + K 
\end{equation}
where ${r_t}^i$ is the individual agent's reward at time $t$ for $i^{th}$ drone. It is computed as:

\begin{equation} 
{r_t}^i=  {\alpha}({\| \mathbf{p_{t-1}}\|}-{\| \mathbf{p_{t}}\|}) +\beta \|\mathbf{v_t}\|+c
\end{equation}
where $\alpha$ is the position-shaping factor determining whether the drone is approaching or moving away from its target, a positive reward is given for proximity to the target, while a negative reward is assigned for moving farther away. $\beta$ scales the velocity penalty, penalizing high speeds to ensure a safe landing and $c$ is the constant reward for encouraging a successful landing.
$r^{c}$ accounts for collision penalties, calculated as:
\begin{equation}  
r^{c} =
\begin{cases}
-\alpha_{c} & \text{if collision}, \\
0 & \text{else}
\end{cases}
\end{equation}

If drones approach each other within minimal collision distances, a negative penalty $\alpha_{c}$ is applied to discourage collisions; otherwise, no penalty is imposed. $K$ is the final reward granted when all agents successfully land on the target simultaneously.

\subsection{Model Architecture}
\begin{figure}[h]
 \centering
 \includegraphics[width=0.5\textwidth]{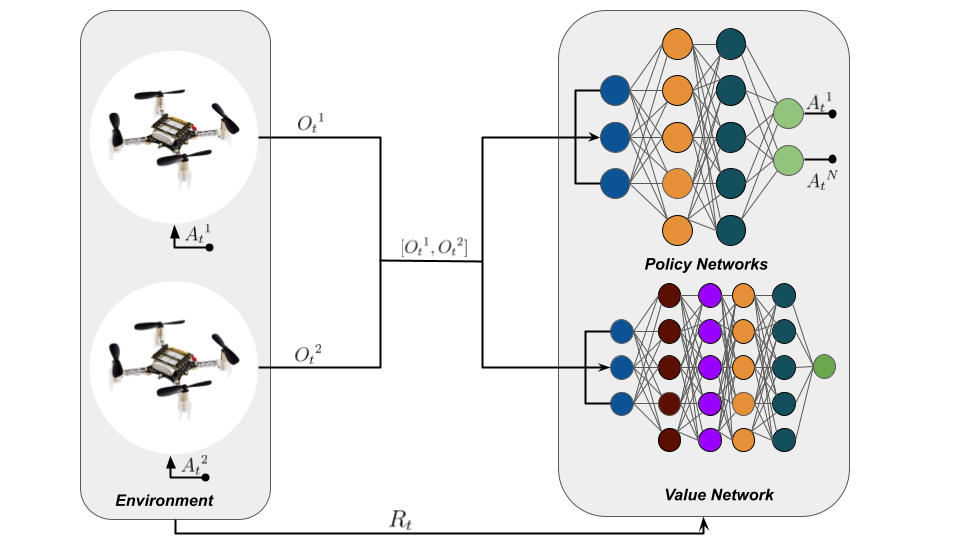}
 \caption{A neural network architecture for PPO algorithm}
 \label{fig:arc}
\end{figure}

We employed a Proximal Policy Optimization (PPO) algorithm with a neural network architecture depicted in Fig. \ref{fig:arc}. The input layer of the neural network receives an array of data from two drone observations, which is subsequently flattened into a 1D array with the assistance of a vector environment from Gymnasium. This flattened array is then fed as input to the policy network, where the features are processed through a series of fully connected layers. These layers incorporate Rectified Linear Unit (ReLU) activation functions and are structured with dimensions of 512 x 2, 256, and 128. The output is fed into six neurons, where the first neuron determines the action for the first drone, and the remaining neurons correspond to the second drone. The output is utilized to calculate the next target position of the drone.
\subsection{Simulation Setup}
\begin{figure}[h]
    \centering
    \includegraphics[width=0.4\textwidth]{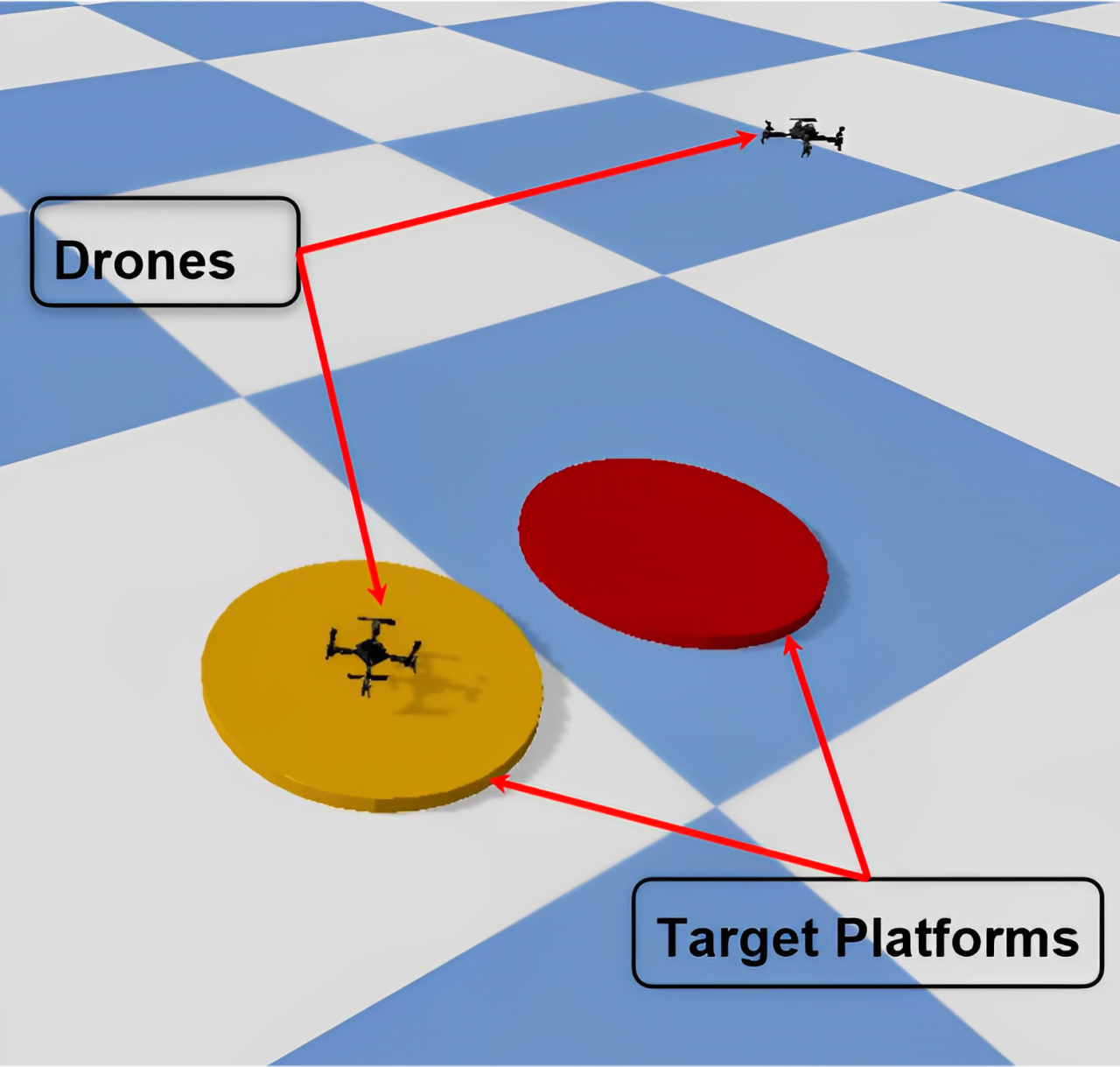}
    \caption{Gym PyBullet environment for simulating a MADRL-driven swarm of drones}
    \label{fig:gym}
\end{figure}

Gym PyBullet environment \cite{c1} is employed to simulate and train a customized drone swarm along with individual target platforms depicted in Fig. \ref{fig:gym}. The environment is vectorized and adapted for multi-agent requirements by integrating with the Stable Baselines3 RL framework.

This environment offers a versatile platform for modeling complex dynamics and interactions within the simulated world. Leveraging PyBullet's physics engine, it accurately captures the intricate behaviors of drones and their interactions with the environment.

\subsection{Training Configuration} 
We employed the Proximal Policy Optimization (PPO) algorithm due to its high performance and sample efficiency where the training is carried out by randomly placing the drones and the target platform within 3D spaces of 4 x 4 x 4 m. During each time step, both the drones and the target platform were repositioned randomly to new locations.
\begin{figure}[h]
    \centering
    \includegraphics[width=0.5\textwidth,height=0.3\textwidth]{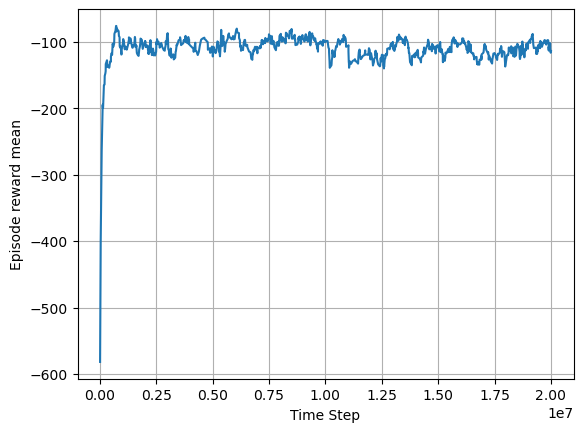}
    \caption{Episode reward mean during training }
    \label{fig:rew}
\end{figure}

\begin{figure}[h]
    \centering
    \includegraphics[width=0.5\textwidth,height=0.3\textwidth]{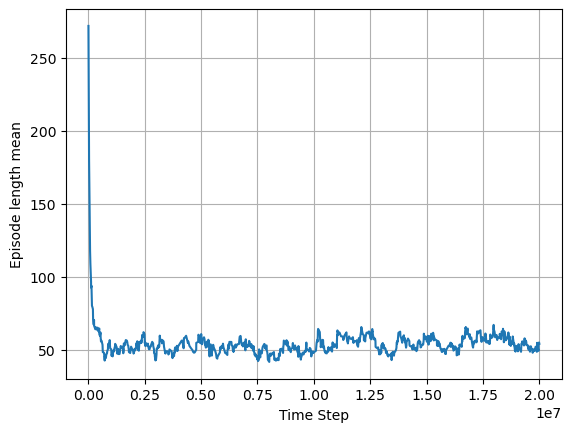}
    \caption{Episode length mean during training }
    \label{fig:ep}
\end{figure}

The training process has been carried out using the supercomputer with resources equipped with 36 CPU cores and 1 GPU core for 20 million time steps. The results reveal that the mean reward converges over time shown in Fig. \ref{fig:rew} while the average episode length reduces shown in Fig. \ref{fig:ep} indicating the effectiveness of the training process.

\section{Experiments}
We experimented to validate our proposed solution deployed via ROS, as discussed in the following section. Our experiments aim to investigate drone landing accuracy and compare it to other approaches used for drone landing.

\subsection{Experimental Setup}
\begin{figure}[ht]
\centering
\includegraphics[width=0.5\textwidth]{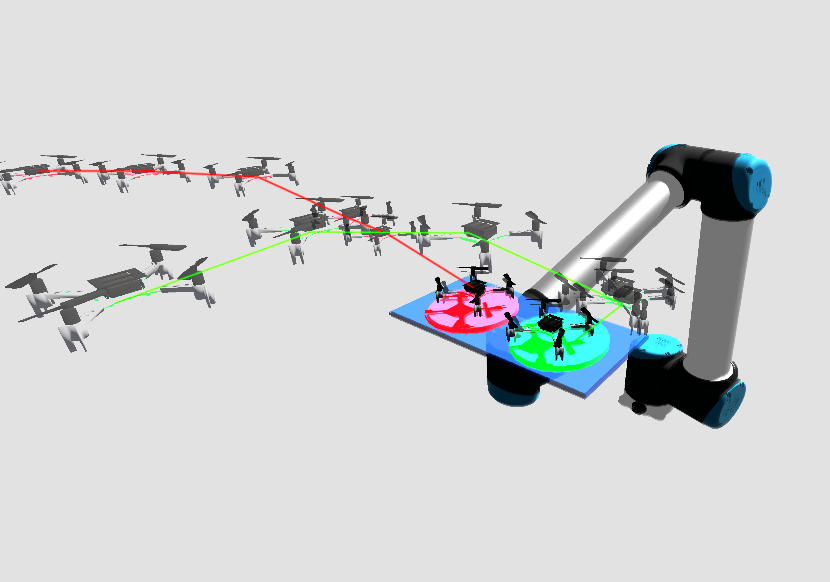}
\caption{MARLander experimental setup for multiagent drone landing in different conditions of the landing pads.}
 \label{fig:exp}
\end{figure}

We developed a practical testing framework to validate our proposed solution across a spectrum of test cases, employing various compression approaches tailored to our solution. Initially, we created a 0.5 x 0.5 acrylic landing pad featuring two cylindrical pads, each with a radius of 0.2 m and a thickness of 2.5 mm, tailored for the two Bitcraze nano quadrotors. This landing pad was affixed to the Tool Center Point (TCP) of a UR10 robotic arm, as depicted in Fig. \ref{fig:exp}.

\begin{figure*}
 \centering
\begin{subfigure}[b]{0.3\textwidth}
\includegraphics[width=\textwidth]{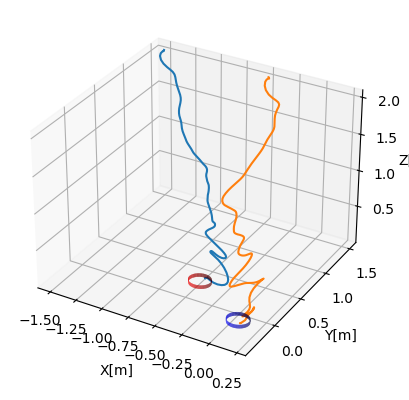}
\caption{}
\end{subfigure}
\begin{subfigure}[b]{0.3\textwidth}
\includegraphics[width=\textwidth]{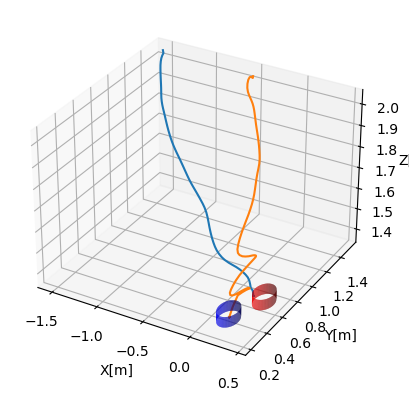}
\caption{}
\end{subfigure}
\begin{subfigure}[b]{0.3\textwidth}
\includegraphics[width=\textwidth]{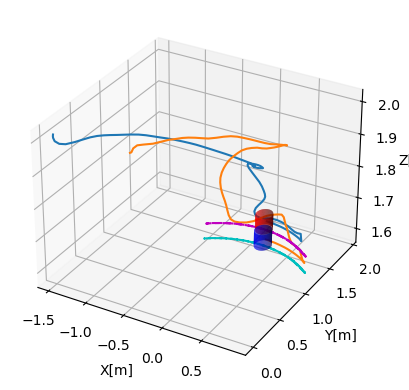}
\caption{}
\end{subfigure}
\caption{Trajectories of the first drone (light blue), second drone (orange), target platform for the first drone (red), and the second drone (blue) landing on static platforms: (a) on the floor, (b) on a UR10 robotic arm, and (c) on a UR10 arm performing linear movement with constant velocity where the first platform (magenta) and the second platform (cyan) represent trajectories while moving.}
\label{fig:landingpads1}
 \end{figure*}

\subsection{MARLander experiment with  stationary platform}

\subsubsection{Description} 
In this experiment, a drone is randomly initialized, and its actions are predicted by a trained policy that processes incoming data from Vicon and broadcasts the corresponding control outputs to each drone. This process is repeated 12 times where half of the experiment is conducted when the landing pad is placed on the floor and the rest is equipped on TCP of UR10 manipulator, with the collected data analyzed to evaluate the trained model's effectiveness based on the drone landing success rate and accuracy metrics.
 \subsubsection{Result and Discussion} 
The landing precision was assessed over 12 experiments, showcasing mean deviations of 2.67 cm and 3.26 cm for the drones concerning their intended targets. The average landing error was calculated to be 2.965 cm, with a success rate of 91.67\%. These results indicate that the trained policy effectively guided the swarm of drones to land on the platform with remarkable precision.

\subsection{MARLander experiment with moving platform}
\subsubsection{Description} 
In this experiment, a UR10 robot manipulator is directed to move from its base within the angular range of $-\pi/2$ to $\pi/2$, with a linear velocity spanning from 0.2 to 0.5 m/s. The drone is randomly initialized, and its actions are anticipated through trained policy to determine the control outputs of each drone. This entire process is iterated 8 times across various velocities. The gathered data is then scrutinized to assess the efficacy of the trained model, relying on predefined metrics.
 \subsubsection{Result and Discussion} 
The experiment involved assessing landing precision through 8 trials and presenting mean deviations of 4.38 cm and 3.47 cm for the drones in relation to their designated targets. The average landing error was found to be  3.93 cm, with an average success rate of 75 \%. Interestingly, it was observed that as the speed of the manipulator increased, the success rate of the experiment decreased, and the mean average landing accuracy was reduced. This suggests that while the trained policy was able to successfully guide a swarm of drones to land on the platform, it faced challenges when dealing with high-speed manipulator movements.

\subsection{Baselines Experiment}

\subsubsection{Description} 
In this experiment, we utilized a PID controller alongside an APF planner to direct the drones toward their designated target locations.

\subsubsection{Result and Discussion} 
The results of the experiment are summarized as follows: the mean landing precision was assessed across 10 experiments, revealing an average landing error of 4.8 cm on the statics platform with a success rate of 80\% 7.4 cm with a success rate of 60\%.
\subsection{Additional Baselines Experiment}

\subsubsection{Description} 
In this experiment, a trained single-agent RL (Reinforcement Learning) agent, adept at navigating to a goal point, with an APF (Artificial Potential Field) planner proposed by the authors \cite{b12} was used as a comparison to our developed solution.

\subsection{Single-Agent and Multi-Agent DRL Landing Comparison}

Our experimentation shows that, proposed multi-agent reinforcement learning demonstrates acceptable accuracy with a high success rate. Notably, single-agent-based landing also achieved precise landings. However, our system outperforms in terms of landing time and collision avoidance compared to the other approach, although this aspect is not explicitly mentioned in \cite{b12}.

\begin{table}[!htb]
    \centering
    \caption{Comparison of the experiments for the static platform.}
    \begin{tabular}{cccc}

    \toprule
         & Success rate (\%) & Precision (cm) &time (s)\\ 
        \midrule
     Baseline & $80.0$ & $4.8$ & $24$  \\
     Morpholander & $-$ & $2.35$ & $-$ \\
     \rowcolor{Gray}
     MARLander  & $91.67$ & $2.26$ & $12$ \\
      \bottomrule
    \end{tabular}
    \label{tab:comp}
\end{table}

\begin{table}[!htb]
    \centering
    \caption{Comparison of the experiments for the moving platform.}
    \begin{tabular}{cccc}
    \toprule
         & Success rate (\%) & Precision (cm) &time (s)\\ 
        \midrule
     Baseline & $60.0$ & $7.4$ & $28$  \\
     Morpholander & $-$ & $3.5$ & $-$ \\
     \rowcolor{Gray}
     MARLander  & $75$ & $3.93$ & $17$ \\
      \bottomrule
    \end{tabular}
    \label{tab:comp}
\end{table}

\section{Conclusion and Future Work}
In our research, we developed a swarm of drone landing systems using a novel MADRL approach. Through a series of experiments, we successfully demonstrated that our developed solution can effectively navigate a swarm of drones to their target landing position. Our results show a remarkable success rate of over 90\% on static targets and 75\% on moving targets, with acceptable landing accuracy. 

Furthermore, we conducted comparative analyses with a baseline experiment employing PID controllers and APF planners. The results indicate the superior performance of our MARLander.

In the future, our focus will be on enhancing the scalability of the MARLander approach. Additionally, we plan to extend the experiments on dynamic platforms to showcase the system's capability to land on moving and inclined surfaces with a high-speed manipulation operation.

\end{document}